\documentclass[12pt]{article}
\usepackage{amsmath}
\usepackage{amsfonts}
\usepackage{amssymb}
\usepackage{graphicx}
\usepackage{geometry} 
\usepackage{hyperref}
\usepackage{caption}
\usepackage{algorithm}
\usepackage{algorithmicx}
\usepackage{algpseudocode}
\usepackage{booktabs}
\usepackage{multicol}
\usepackage{array, longtable}
\usepackage{tikz}
\usetikzlibrary{positioning,shapes,arrows,shadows}
\usepackage{tcolorbox} 

\title{\large\textbf{GPT-HTree: A Decision Tree Framework Integrating Hierarchical Clustering and Large Language Models for Explainable Classification}}
\author{
    \small Te Pei$^{1}$, Fuat Alican$^{2}$, Aaron Ontoyin Yin$^{2}$, Yiğit Ihlamur$^{2}$ \\[0.5em]
    \small $^{1}$University of Oxford \hspace{1em} $^{2}$Vela Research \\[0.5em]
}

\date{\small January 2025}

\begin{document}

\maketitle

\begin{abstract}
Traditional decision trees often fail on heterogeneous datasets, overlooking differences among diverse user segments. This paper introduces \textbf{GPT-HTree}, a framework combining hierarchical clustering, decision trees, and large language models (LLMs) to address this challenge. By leveraging hierarchical clustering to segment individuals based on salient features, resampling techniques to balance class distributions, and decision trees to tailor classification paths within each cluster, GPT-HTree ensures both accuracy and interpretability. LLMs enhance the framework by generating human-readable cluster descriptions, bridging quantitative analysis with actionable insights. Applied to venture capital, where the random success rate of startups is 1.9\%, GPT-HTree identifies explainable clusters with success probabilities up to 9 times higher. One such cluster, \textit{serial-exit founders}, comprises entrepreneurs with a track record of successful acquisitions or IPOs is 22x more likely to succeed compared to \textit{early professionals}. 
\end{abstract}

\section{Introduction}
Decision trees are fundamental tools in machine learning (ML), prized for their interpretability and simplicity in classification tasks. By providing clear decision paths, they enable users to understand and trust the reasoning behind predictions. However, their effectiveness diminishes when applied to heterogeneous datasets comprising entities with varying characteristics. Uniform decision paths often fail to account for the nuanced differences among diverse segments, leading to oversimplified or misleading classifications.

Unsupervised clustering methods, on the other hand, excel in discovering latent structures within complex datasets. These methods, including hierarchical clustering, k-means, and DBSCAN, are powerful tools for segmenting populations into meaningful clusters without requiring predefined labels. While they are effective for uncovering hidden patterns, their primary drawback is a lack of explainability. Clusters produced by unsupervised methods often lack intuitive descriptions or actionable insights, making it difficult to interpret their relevance or apply them in practical decision-making scenarios.

\subsection{Motivation}
The primary motivation for this work stems from three key observations:
\begin{itemize}
    \item Traditional decision trees struggle with heterogeneous populations, often producing oversimplified or misleading classifications.
    \item Current approaches lack mechanisms to adapt decision paths to the unique characteristics of different subgroups within a dataset, leading to generalized rules that fail to capture the nuanced differences among diverse populations.
    \item There is an increasing demand for classification methods that balance interpretability and accuracy, particularly in handling diverse populations. Leveraging the reasoning capabilities of LLMs present a potential solution to address this need.  
\end{itemize}

Explainability is particularly critical in high-stakes domains such as venture capital, healthcare, and criminal justice. Supervised learning methods offer interpretability but rely on exhaustive labeled data and predefined categories. This reliance limits their ability to explore unknown or dynamic structures in data. In contrast, unsupervised methods are powerful for uncovering novel patterns but fail to provide the transparency required for practical applications. Achieving explainable cluster creation is essential for making informed decisions and deriving actionable insights, particularly in venture capital.

\subsection{Empirical Context and Challenges in Venture Capital}
The domain of venture capital provides a compelling empirical study for the importance of explainable unsupervised clustering. At the inception stage, the probability of a startup achieving random success---defined as becoming a unicorn or reaching significant exit milestones---is approximately 1.9\% \cite{Xiong2024}. However, empirical evidence shows that there are clusters of startups with significantly higher probabilities of success, up to 20\%. Identifying these high-potential clusters is critical for venture capitalists, as it allows them to allocate resources and maximize investment outcomes.

Using purely supervised methods to identify these clusters is impractical due to the infeasibility of exhaustively labeling all potential categories of startups. Moreover, the optimal number and structure of clusters are not known a priori, further limiting the scope of supervised approaches. While unsupervised methods can reveal these high-potential clusters, their lack of explainability prevents actionable decision-making. Venture capitalists require interpretable clusters to focus on the right channels to source the highest probability startups.

To address these challenges, this work introduces \textbf{GPT-HTree}, an advanced framework that integrates hierarchical clustering with decision trees and large language models (LLMs). This hybrid approach leverages unsupervised clustering to uncover latent structures in the data while employing decision trees to provide localized, interpretable decision paths within each cluster. LLMs enhance the framework by generating human-readable descriptions of clusters, bridging the gap between quantitative analysis and qualitative insights.

By enabling explainable cluster creation and classification, GPT-HTree empowers venture capitalists to identify and target the most promising clusters for deal sourcing and investment. This framework demonstrates empirical effectiveness by uncovering clusters that outperform the random success rate by 9x, a significant jump from the baseline probability of 1.9\% to 17.4\% and providing actionable insights into the determinants of entrepreneurial success. In doing so, GPT-HTree not only addresses the limitations of traditional decision trees and unsupervised methods but also sets a new direction for interpretable ML by integrating LLMs into the process. 

\subsection{Key Contributions}
Our work makes the following contributions:
\begin{itemize}
    \item A decision tree framework that integrates hierarchical clustering and LLMs to create personalized classification paths for different user segments
    \item A new method using LLMs to automatically generate interpretable persona descriptions from quantitative cluster features, enabling human-interpretablity
    \item An end-to-end pipeline that can take new founder profiles as input and generate both interpretable persona descriptions and probability estimates of success
\end{itemize}

\section{Related Work}

There has been a growing body of research exploring the integration of ML and LLMs, a rapidly expanding area that holds the potential to advance the field toward artificial superintelligence (ASI). This emerging paradigm leverages LLMs for their reasoning and generative capabilities to perform ML processes such as data wrangling, feature modeling, and parameter optimization—tasks that have traditionally relied on human expertise. By automating these processes, LLMs not only reduce manual effort but also open new possibilities for more efficient and scalable workflows. Our work builds upon these advancements by demonstrating how LLMs can guide and enhance these traditionally human-driven tasks within the context of decision tree and clustering frameworks.

\begin{figure}[ht]
    \centering
    \includegraphics[width=\textwidth]{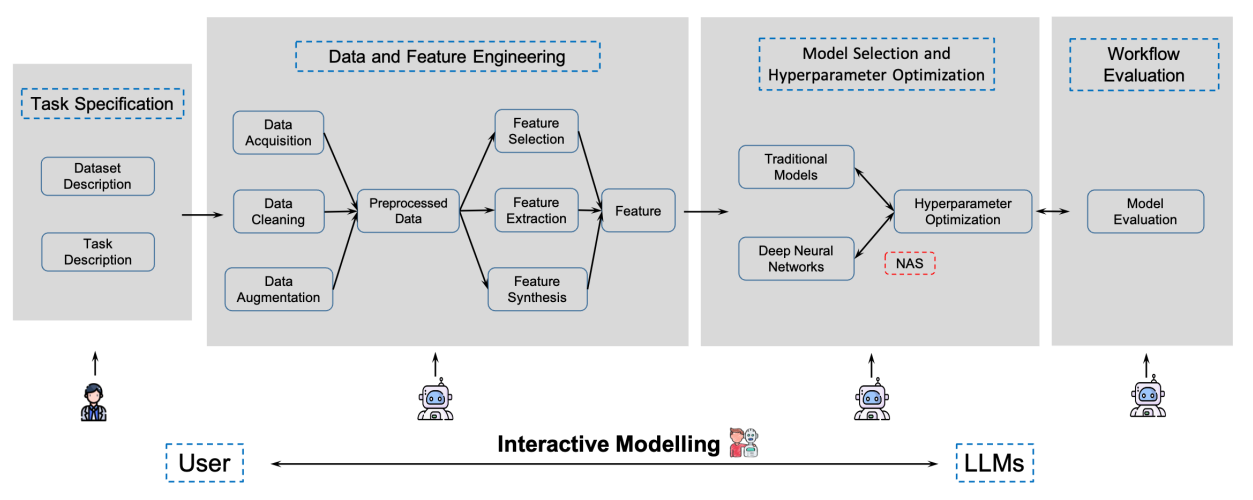}
    \caption{An overview of the Machine Learning Workflow, where task specification serves as the input, encompassing
key stages of data and feature engineering, model selection and hyperparameter optimization, and workflow evaluation. Figure reproduced from \cite{gu2024large}.}
    \label{fig:LLMworkflow}
\end{figure}

\subsection{Decision Trees, Ensemble Methods, and Their Limitations}
Decision trees and their ensemble variants (e.g., Random Forests, Gradient Boosting) have long been favored for their interpretability and effectiveness in structured, relatively uniform datasets \cite{Duda2001,Durkin1992,Fayyad1992,Breiman1993,Quinlan1993,Quinlan1996,Brodley1995}. Despite their success in fields like text mining \cite{Yang1997}, speech processing \cite{Riley1989,Chien2002}, bioinformatics \cite{Salzberg1998}, and web intelligence \cite{Zamir1998,Cho2002}, these methods face growing challenges as data becomes increasingly large, heterogeneous, and multimodal. Scaling such models often involves labor-intensive steps—feature engineering, hyperparameter tuning, and iterative refinement—making it difficult to adapt swiftly to complex, evolving task requirements.

\subsection{Clustering-Based Classification and Hybrid Approaches}
Clustering-based classification represents an effort to better handle complex data distributions by partitioning instances into meaningful subgroups before applying specialized models within each cluster. This strategy can enhance both model accuracy and interpretability, yet constructing and maintaining such pipelines remains time-consuming and knowledge-intensive. The need to integrate diverse data types and leverage prior human or historical insights is not easily met by traditional workflows, underscoring the limitations of existing tools and techniques. As tasks become more intricate, these challenges highlight opportunities for more advanced, automated solutions—potentially guided by Large Language Models—to streamline the entire ML workflow.

\subsection{Large Language Models in Machine Learning}
Recent advances in Large Language Models (LLMs) \cite{Hollmann2024, Wang2024a, Achiam2023, Touvron2023, Hu2024, Tai2024, Luo2024} have opened new possibilities for designing and optimizing ML workflows (Fig.~\ref{fig:LLMworkflow}). By leveraging extensive pre-trained knowledge, LLMs reduce manual overhead in tasks like feature engineering, model selection, and hyperparameter tuning \cite{Gu2023, Klievtsova2023, Zhang2023a, Xiao2024, Hong2024}, offering more adaptive and interpretable solutions than traditional AutoML systems \cite{Lazebnik2022, Nikitin2022, An2023}. Beyond task automation, LLMs facilitate human-AI collaboration by interpreting natural language instructions and suggesting workflow refinements interactively \cite{Dakhel2023, ArteagaGarcia2024}. This enhances interpretability and transparency, mitigating the “black-box” critique of complex models \cite{Zhang2023b, Nam2024, Zhang2024a}.

Nonetheless, practical issues remain. Reasoning errors, ethical considerations, and high computational costs pose ongoing challenges \cite{Bommasani2021, Hollmann2024, Yao2024a, Zhang2023b}, and the full integration of LLMs—spanning data preprocessing through evaluation—still requires more systematic study. Overcoming these hurdles and fully tapping into LLM-driven workflows represent a key frontier for future research.

\section{Methodology}
The framework begins with data preprocessing with a resampling technique, Conditional Tabular Generative Adversarial Networks (CTGAN), to balance class distributions. Hierarchical clustering is then applied to uncover groups based on feature similarities, ensuring the model captures the diversity of the dataset. Subsequently, decision trees are employed within each cluster to provide localized, explainable classification rules tailored to each segment. Finally, LLMs are integrated to generate interpretable persona descriptions for each cluster, bridging the gap between statistical patterns and actionable insights. The sections below detail each component of the methodology.

\subsection{Data Preprocessing}

To ensure model performance and mitigate class imbalance, we explored both traditional resampling methods (e.g., random over-sampling, SMOTE) and generative approaches. While conventional techniques can increase minority class representation, they often do so by simple replication or linear interpolation, limiting the diversity of synthesized samples. In contrast, we employ Conditional Tabular Generative Adversarial Networks (CTGAN) \cite{xu2019modeling}, which have demonstrated remarkable efficacy in generating realistic and varied synthetic data. By producing more representative samples for underrepresented classes, CTGAN helps address the skewed distribution in the dataset, thereby enabling the model to learn richer decision boundaries. This ultimately improves its generalization and classification accuracy.

\subsection{Resampling Analysis}
Tables 1 and 2 present the success rate distributions before and after resampling for clusters with high VC experience. The resampling strategy significantly enhanced the model's discriminative power in identifying successful founders.

In the initial distribution (Table 1), the normalized success rates ranged from 4.9\% to 19.75\%, with a relatively compressed distribution. After resampling (Table 4), this range expanded notably from 3.9\% to 50.0\%, creating more distinct separation between high and low-performing groups. This enhanced separation is particularly evident in Sub 7.4, where the success rate increased substantially to 50.0\%, making it easier to identify the characteristics of highly successful founders.

The resampling process also improved the granularity of cluster identification. While maintaining the overall pattern where smaller, specialized clusters show higher success rates, the post-resampling distribution provides clearer differentiation between performance tiers. For instance, we can now more clearly distinguish between high-performing clusters (50.0\%), medium-performing clusters (25.0-28.1\%), and lower-performing clusters (3.9\%), enabling more precise founder profiling and success prediction. 

This increased spread in normalized success rates not only facilitates better identification of high-potential founders but also provides more reliable signals for feature importance analysis and model interpretation.
\begin{table}[h]
\centering
\begin{tabular}{lccccc}
\toprule
Cluster & Total Count & Success Count & Normalized Success Rate \\
\midrule
Cluster  & 103 & 48 & 8.7\%  \\
Sub S.1 & 18 & 17 & 19.75\% \\
Sub S.2 & 16 & 11 & 14.4\%  \\
Sub S.3 & 16 & 9 & 11.8\% \\
Sub S.4 & 53 & 11 & 4.9\%  \\
\bottomrule
\end{tabular}
\caption{Top VC experience cluster before resampling}
\label{tab:initial_feature_importance}
\end{table}

\begin{table}[h]
\centering
\begin{tabular}{lccccc}
\toprule
Cluster & Total Count & Success Count & Normalized Success Rate \\
\midrule
Cluster  & 355 & 44 & 15.1\% \\
Sub S.1 & 34 & 17 & 50.0\%  \\
Sub S.2 & 32 & 9 & 28.1\% \\
Sub S.3 & 36 & 9 & 25.0\% \\
Sub S.4 & 253 & 9 & 3.9\% \\
\bottomrule
\end{tabular}
\caption{Top VC experience cluster after resampling}
\label{tab:feature_importance}
\end{table}

\subsection{Hierarchical Clustering}

After resampling, the initial phase of GPT-HTree involves hierarchical clustering to segment the dataset into meaningful clusters based on feature similarities. As an unsupervised learning technique, hierarchical clustering naturally groups individuals into similar clusters based on their feature profiles, without relying on predefined labels or the need to specify the number of clusters in advance. This method iteratively merges or splits clusters to uncover nested structures within the data. By capturing the diversity among entrepreneurs, hierarchical clustering helps reveal groups that serve as a foundation for subsequent modeling steps.

\subsection{Cluster Analysis and LLM Interpretation}
Following the application of hierarchical clustering, the resulting groups are unlabeled and must be interpreted based on their underlying feature distributions. By computing the standardized z-scores of each feature within each cluster, we can identify the key characteristics that set these groups apart. The specific methods and insights gained from this approach are further detailed in Section 4.4. 

Table 3 summarizes Table 1, incorporating LLM-generated descriptions derived from the z-scores of features within each cluster.

\begin{table}[h]
\centering
\small
\begin{tabular}{p{1.5cm}|p{2cm}|p{2cm}|p{2cm}|p{4cm}}
\toprule
Cluster Level & Cluster ID & Total Count & Success Rate & Key Characteristics \\
\midrule
Main & Cluster 1 & 103 & 46.6\% & Elite entrepreneurial founders with VC-backed success \\
\hline
Sub & S.1 & 18 & 94.4\% & High-performing CEOs with exceptional fundraising \\
\hline
Sub & S.2 & 16 & 68.8\% & Big-tech experienced founders with proven track records \\
\hline
Sub & S.3 & 16 & 56.2\% & Strategic core-team focused startup leaders \\
\hline
Sub & S.4 & 53 & 20.8\% & Independent entrepreneurs with moderate success \\
\bottomrule
\end{tabular}
\caption{Cluster Characteristics and LLM-Generated Descriptions}
\label{tab:cluster_characteristics}
\end{table}

\subsection{Persona Characterization with Large Language Models}
\begin{figure}[ht]
    \centering
    \includegraphics[width=\textwidth]{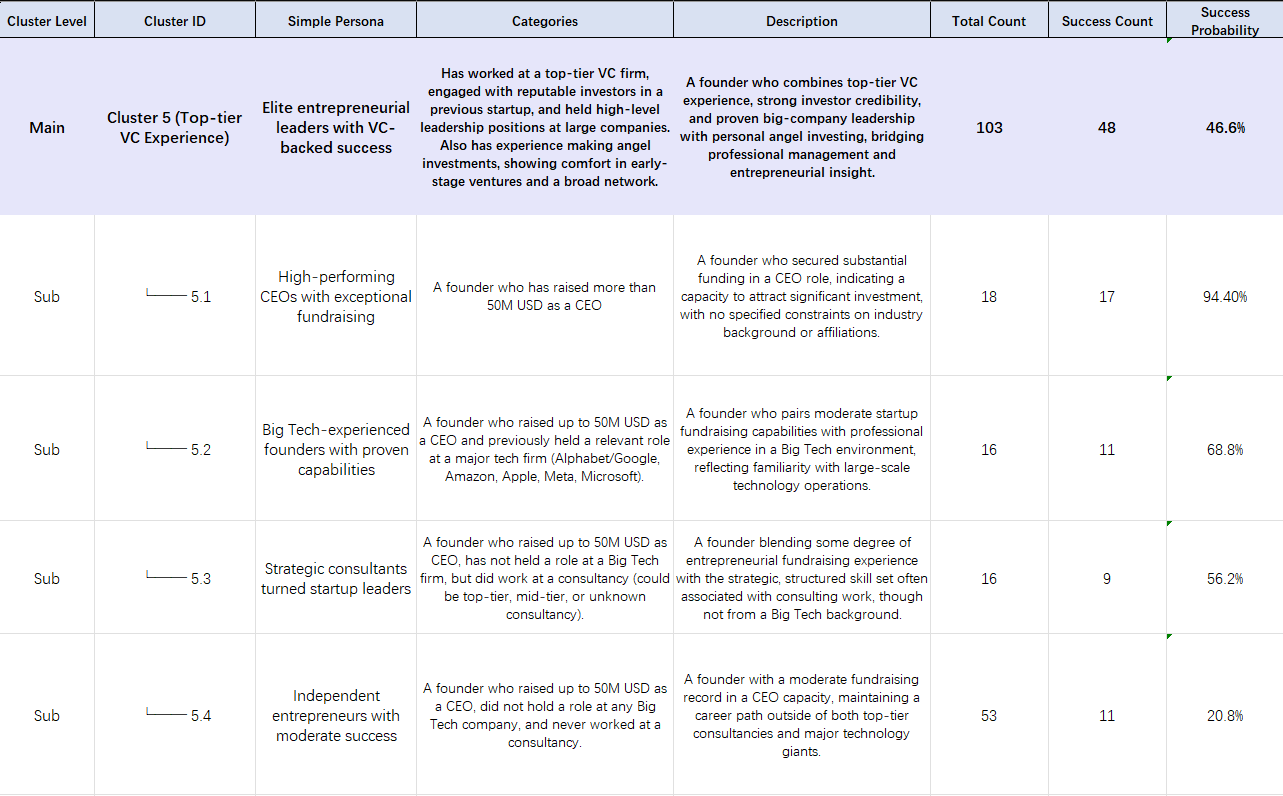}
    \caption{Example of LLM-Generated Persona Descriptions.}
    \label{fig:llm_description}
\end{figure}

Leveraging the identified key features, we utilize large language models (LLMs) to summarize and articulate the distinctive characteristics of each cluster. For example, a cluster characterized by high levels of venture capital (VC) experience and prior startup investments might be summarized by the LLM as "Extremely strong VC experience, prior startups backed by top VCs, and significant leadership experience." This step translates quantitative feature deviations into qualitative persona descriptions, facilitating a deeper understanding of each cluster's unique attributes. An example of the LLM-generated persona descriptions is shown in Figure~\ref{fig:llm_description}. (Table 3 as a simple example)

\subsection{Decision Tree Classification within Clusters}
Following persona characterization, we apply decision tree algorithms within each main cluster to further classify individuals based on their likelihood of success. This approach allows the decision trees to focus on feature interactions that are most relevant within the specific context of each persona category. We experiment with different impurity measures, such as Gini impurity and entropy, to construct the decision trees and compute feature importance scores. This analysis reveals the primary determinants of success within each persona group, providing actionable insights tailored to each segment.

For each feature in the decision tree, we calculate its importance score based on the weighted improvement in the split criterion (Gini impurity) across all nodes where that feature is used for splitting. 

When splitting at node $t$, we select the optimal split by minimizing the Gini impurity:

\[
G(t) = 1 - \sum_{i=1}^{c} p_i^2
\]

where:
\begin{itemize}
    \item $c$: number of classes
    \item $p_i$: proportion of class $i$ samples in node $t$
\end{itemize}

For each candidate split, calculate the weighted impurity after splitting:

\[
G_{\text{split}}(t, f, \theta) = \frac{N_{\text{left}}}{N_t} G(t_{\text{left}}) + \frac{N_{\text{right}}}{N_t} G(t_{\text{right}})
\]

\subsection{LLM Integration and Prompt Design}
For human interpretability, we use GPT-4 as our large language model. The prompt design follows a three-part structure: (1) context setting, (2) feature analysis instruction, and (3) output format specification. Figure \ref{fig:prompt-design} illustrates our prompt template. 

\begin{figure}[htbp]
\centering
\begin{tcolorbox}[colback=gray!10, colframe=black!50, title=Prompt Template for Persona Description Generation, fonttitle=\bfseries]
\textbf{Context:} \\
You are an expert startup analyst specializing in founder behavior analysis. You have been given statistical data about a distinct group of founders identified through cluster analysis.

\vspace{1em}
\textbf{Input Features:} \\
\{feature\_data\}

\vspace{1em}
\textbf{Task:} \\
Analyze these statistical patterns and generate a comprehensive founder persona description that:
\begin{enumerate}
    \item Identifies the key characteristics that distinguish this group
    \item Explains how these characteristics might influence startup success
    \item Provides actionable insights for investors or mentors working with this type of founder
\end{enumerate}

\vspace{1em}
\textbf{Output Format:}
\begin{enumerate}
    \item Persona Summary (2-3 sentences)
    \item Key Distinguishing Traits (bullet points)
    \item Success Factors (bullet points)
    \item Risk Factors (bullet points)
    \item Recommendations (bullet points)
\end{enumerate}

\vspace{1em}
Keep the language professional but accessible. Focus on actionable insights rather than just statistical descriptions.
\end{tcolorbox}
\caption{LLM Prompt Design for Founder Persona Generation}
\label{fig:prompt-design}
\end{figure}

\subsection{Algorithm Design}
The GPT-HTree framework consists of three main components: (1) the training algorithm for building the hierarchical persona-based model, (2) the classification algorithm for predicting new founder outcomes, and (3) the LLM-based persona description generation algorithm for creating interpretable persona descriptions.

\subsubsection{Main Training Process}
Algorithm \ref{alg:gphtree-train} shows the training process, which includes data preprocessing, hierarchical clustering for persona identification, and decision tree construction for each cluster. For each identified cluster, the algorithm builds a specialized decision tree and generates interpretable descriptions using the LLM component.

\begin{algorithm}[htbp]
\caption{GPT-HTree Training Algorithm}
\label{alg:gphtree-train}
\begin{algorithmic}[1]

\Require Dataset $D$ with features $X$ and success labels $Y$
\Require Parameters: $n\_main\_clusters$, $min\_subcluster\_size$, $real\_world\_success\_rate$
\Ensure Hierarchical clustering with personalized decision trees

\State \textbf{// Phase 1: Data Preprocessing and Resampling}
\State $D' \gets \text{Preprocess}(D)$
\State $D_{\text{balanced}} \gets \text{ResampleData}(D',\ real\_world\_success\_rate)$

\State \textbf{// Phase 2: Main Cluster Creation}
\State $main\_clusters \gets \text{HierarchicalClustering}(D_{\text{balanced}},\ n\_main\_clusters)$

\State \textbf{// Phase 3: Subcluster Analysis}
\For{$C$ \textbf{in} $main\_clusters$}
    \If{$\text{size}(C) \leq min\_subcluster\_size$}
        \State \textbf{continue}
    \EndIf
    
    \State \textbf{// Build decision tree for cluster}
    \State $tree \gets \text{DecisionTree}(C,\ min\_subcluster\_size,\ \text{max\_depth} = 3)$
    
    \State \textbf{// Extract rules and characteristics}
    \State $rules \gets \text{ExtractRules}(tree)$
    \State $features \gets \text{CalculateFeatureImportance}(tree)$
    
    \State \textbf{// Generate LLM descriptions using GPT-4}
    \State $feature\_data \gets \text{FormatFeatureStats}(features)$
    \State $prompt \gets \text{ConstructPrompt}(feature\_data)$
    \State $descriptions \gets \text{QueryGPT4}(prompt)$
    \State $processed\_description \gets \text{PostProcessLLMOutput}(descriptions)$

    \State SaveClusterResults($C,\ rules,\ features,\ processed\_description$)
    \State SaveClusterResults($C,\ rules,\ features,\ descriptions$)
\EndFor

\State \Return{$trained\_model$}

\end{algorithmic}
\end{algorithm}

\subsubsection{Founder Classification Process}
Algorithm \ref{alg:gphtree-classify} presents the classification process for new founders. It employs a two-stage approach: first identifying the closest matching persona cluster, then applying the corresponding decision tree to generate predictions. This ensures that success prediction is contextualized within the appropriate founder persona.

\begin{algorithm}[htbp]
\caption{GPT-HTree Classification Algorithm}
\label{alg:gphtree-classify}
\begin{algorithmic}[1]
\Require \textit{Trained GPT-HTree model}, \textit{new founder feature vector} $f$
\Ensure \textit{Classification result} with explanation

\State \textbf{// Phase 1: Find closest main cluster}
\State $distances \gets []$
\For{\textbf{each} $C$ \textbf{in} model.main\_clusters}
    \State $dist \gets \text{EuclideanDistance}(f,\ C.\text{centroid})$
    \State distances.append($dist$)
\EndFor

\State \textbf{// Phase 2: Get cluster prediction}
\State $closest\_cluster \gets \arg\min(\text{distances})$
\State $tree \gets \text{GetClusterTree}(closest\_cluster)$
\State $path \gets \text{GetDecisionPath}(tree,\ f)$

\State \textbf{// Phase 3: Generate prediction and explanation}
\State $prediction \gets tree.\text{predict}(f)$
\State $explanation \gets \text{GenerateExplanation}(path,\ closest\_cluster)$
\State $confidence \gets \text{CalculateConfidence}(path)$

\State \Return \{$\text{cluster}: closest\_cluster,$ 
         $\text{prediction}: prediction,$ 
         $\text{explanation}: explanation,$
         $\text{confidence}: confidence$\}

\end{algorithmic}
\end{algorithm}

\subsubsection{LLM-based Persona Description Generation}
Algorithm \ref{alg:llm-description} details the process of generating interpretable persona descriptions using GPT-4. This algorithm transforms quantitative cluster characteristics into qualitative, actionable insights through four main steps:

\begin{enumerate}
    \item \textbf{Feature statistics formatting:} \\
    Processes and formats significant features for LLM input
    
    \item \textbf{Prompt construction:} \\
    Builds structured prompts following our template design
    
    \item \textbf{GPT-4 querying:} \\
    Interfaces with the GPT-4 model using optimized parameters
    
    \item \textbf{Output processing:} \\
    Ensures the generated descriptions are properly structured and validated
\end{enumerate}

\begin{algorithm}[htbp]
\caption{LLM-based Persona Description Generation}
\label{alg:llm-description}
\begin{algorithmic}[1]
\Require \textit{Cluster features} $F$, \textit{feature importance scores} $I$
\Ensure \textit{Structured persona description}

\Statex \Comment{Format feature statistics}
\Function{FormatFeatureStats}{$F, I$}
    \State $formatted\_data \gets \{\}$
    \For{\textbf{each} $f$ \textbf{in} $F$}
        \If{$I[f] \geq significance\_threshold$}
            \State $stats \gets \text{CalculateStatistics}(f)$
            \State $comparison \gets \text{CompareToGlobalStats}(f)$
            \State $formatted\_data.\text{append}(\text{FormatFeature}(f,\ stats,\ comparison))$
        \EndIf
    \EndFor
    \State \Return $formatted\_data$
\EndFunction

\Statex \Comment{Construct GPT-4 prompt}
\Function{ConstructPrompt}{$formatted\_data$}
    \State $prompt \gets \text{LoadPromptTemplate}()$
    \State $prompt.\text{replace}(\{\text{feature\_data}\},\ formatted\_data)$
    \State $prompt.\text{add}(\text{ContextualGuidelines}())$
    \State \Return $prompt$
\EndFunction

\Statex \Comment{Query GPT-4 and process response}
\Function{QueryGPT4}{$prompt$}
    \State $params \gets \{\}$
    \State $params.\text{temperature} \gets 0.7$
    \State $params.\text{max\_tokens} \gets 1000$
    \State $params.\text{top\_p} \gets 0.95$
    \State $params.\text{frequency\_penalty} \gets 0.5$
    \State $response \gets \text{GPT4.generate}(prompt,\ params)$
    \State \Return $response$
\EndFunction

\Statex \Comment{Post-process LLM output}
\Function{PostProcessLLMOutput}{$response$}
    \State $sections \gets \text{ParseSections}(response)$
    \State $validated \gets \text{ValidateContent}(sections)$
    \State $formatted \gets \text{FormatForStorage}(validated)$
    \State \Return $formatted$
\EndFunction

\end{algorithmic}
\end{algorithm}

This three-part algorithmic framework enables GPT-HTree to not only identify and classify founder personas but also provide human-interpretable insights about each persona group. The LLM component serves as a bridge between statistical patterns and actionable business intelligence, making the model's outputs more valuable for practical decision-making.

\newpage
\section{Empirical Validation: Implementing GPT-HTree in Venture Capital}
The GPT-HTree framework is implemented through the following sequential steps:

\begin{enumerate}
    \item \textbf{Data Resampling:} Balancing the dataset to ensure equitable representation of all classes, as illustrated in Figure~\ref{fig:resampling}.
    
    \item \textbf{Hierarchical Clustering:} Segmenting the dataset into eight distinct persona categories based on significant feature deviations.
    
    \item \textbf{Feature Extraction and Normalization:} Identifying key features within each cluster by calculating z-scores.

    \item \textbf{Decision Tree Classification:} Building and evaluating decision trees within each cluster to determine the critical features influencing success, and calculating feature importance scores. The decision paths and feature importance are then used to generate interpretable descriptions for each subcluster, explaining their unique characteristics and success patterns.
    
    \item \textbf{Subcluster Feature Analysis:} For each subcluster identified within the main clusters, extracting and analyzing distinctive features that characterize the subgroup's success patterns. This enables detailed understanding of success factors within each persona category.

    \item \textbf{LLM-Based Persona Description:} Generating human-readable cluster descriptions using a specialized prompting pipeline and fine-tuning approach:
    \begin{itemize}
        \item Structured feature-to-text prompts that map statistical patterns to natural language
        \item Iterative refinement through few-shot learning with expert-validated descriptions
        \item Domain-specific parameter tuning to ensure relevance and accuracy of generated personas
        \item An example of the generated descriptions for a single cluster is shown in Figure~\ref{fig:llm_description}
    \end{itemize}
\end{enumerate}

\subsection{Data Resampling}
\begin{figure}[ht]
    \centering
    \includegraphics[width=\textwidth]{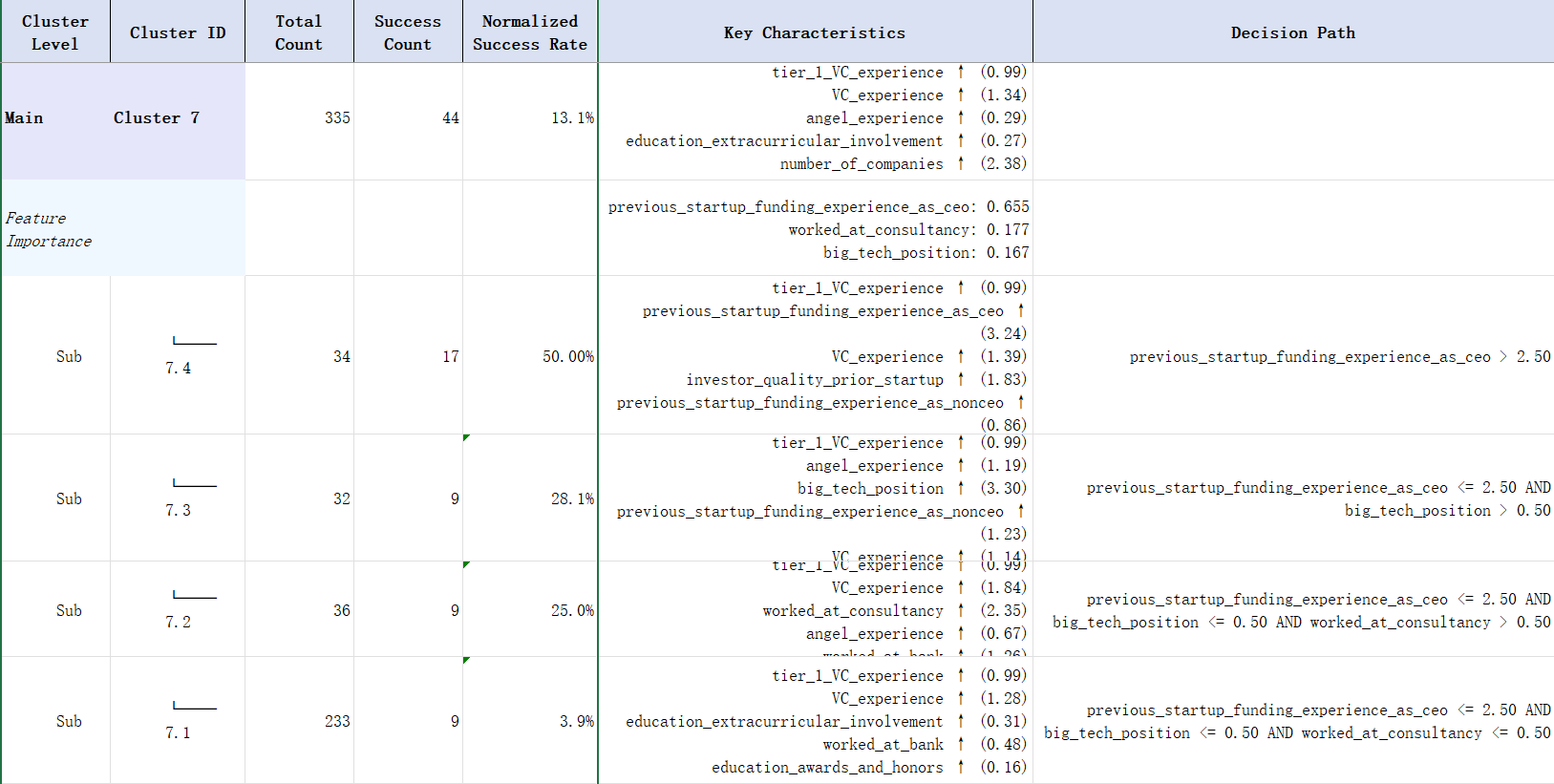}
    \caption{Resampling Techniques and Their Impact on Class Balance.}
    \label{fig:resampling}
\end{figure}

Resampling addresses class imbalance by adjusting the distribution of classes in the dataset. Figure~\ref{fig:resampling} illustrates the advantage of resampling, demonstrating a more balanced representation post-resampling, which enhances the model's ability to learn from minority classes effectively.

\subsection{Hierarchical Clustering}
We perform hierarchical clustering on the resampled data, resulting in eight main clusters:
\begin{enumerate}
    \item Acquisition Experience
    \item IPO Experience
    \item VC Experience
    \item Big-Tech Experience
    \item Media Influencer
    \item Academic Expert
    \item Ordinary Worker
    \item Poor Career Development
\end{enumerate}
Each cluster represents a distinct persona category with unique feature profiles.

\subsection{Feature Extraction and Normalization}
\begin{figure}[ht]
    \centering
    \includegraphics[width=\textwidth]{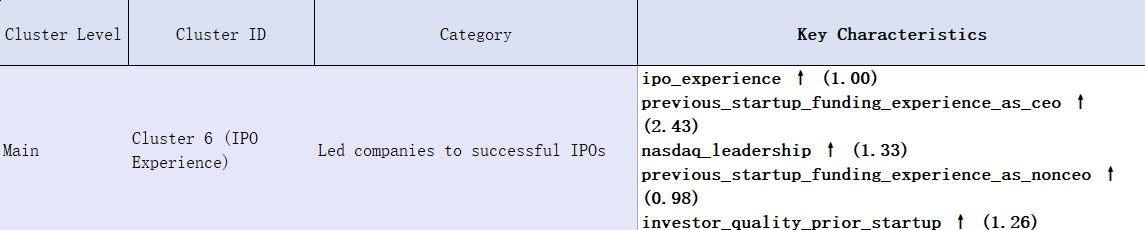}
    \caption{Main Cluster after key characteristics mining}
    \label{fig:keychar}
\end{figure}

For each main cluster, we model the distribution of each feature as a normal distribution and compute the z-scores to assess their deviation from the overall data set. Features with the highest standard deviations within a cluster are deemed key features. For instance, as shown in Figure~\ref{fig:keychar}, in the IPO experience cluster, features such as \texttt{ipo\_experience} and \texttt{nasdaq\_experience} exhibit the highest z-scores, indicating their significance in differentiating this cluster.

\newpage
\subsection{LLM Description of a Main Cluster}
Using the identified key features, we prompt an LLM to generate descriptive summaries for each cluster. For example:

\begin{verbatim}
tier_1_VC_experience ↑ (0.99)
VC_experience ↑ (1.41)
investor_quality_prior_startup ↑ (0.60)
big_leadership ↑ (0.53)
angel_experience ↑ (0.28)
\end{verbatim}

\textbf{LLM Summary:} "Extremely strong VC experience, prior startups backed by top VCs, and significant leadership experience."

Figure~\ref{fig:llm_description} showcases examples of LLM-generated persona descriptions for all four sub clusters under this main cluster.

\subsection{Decision Tree Classification}
\begin{figure}[ht]
    \centering
    \includegraphics[width=\textwidth]{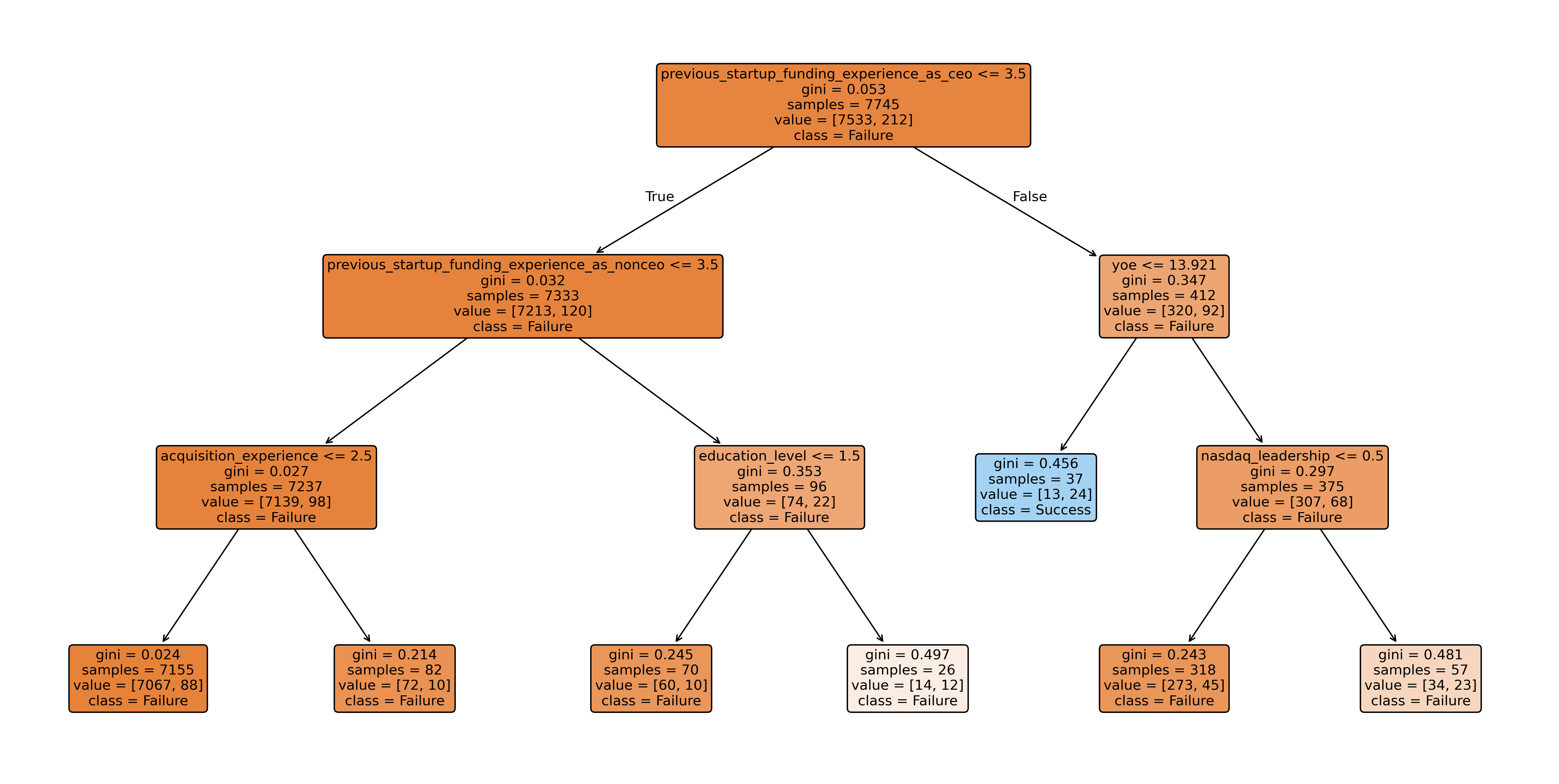}
    \caption{Resampling Techniques and Their Impact on Class Balance.}
    \label{fig:decision_tree_cluster_3}
\end{figure}

Within each cluster, we construct decision trees using different impurity measures (Gini impurity and entropy) to classify individuals based on their likelihood of entrepreneurial success. Feature importance is calculated to identify which features most significantly influence the classification outcome. This localized decision tree analysis uncovers the primary determinants of success specific to each persona category. Please see Figure~\ref{fig:decision_tree_cluster_3} as an example. 

After generating each decision tree, we calculate the importance of each feature within the tree to assess its impact on the classification results. The calculation of feature importance is based on the contribution of the feature at the decision tree's split nodes.

For feature $i$, its importance score is calculated as:

\[
\text{Importance}(i) = \sum_{j \in \text{splits on } i} \frac{N_j}{N_{\text{total}}} \cdot \Delta G(j)
\]

where the split gain $\Delta G(j)$ is calculated as:

\[
\Delta G(j) = G(t) - \left(\frac{N_{\text{left}}}{N_j} \cdot G(t_{\text{left}}) + \frac{N_{\text{right}}}{N_j} \cdot G(t_{\text{right}})\right)
\]

where:
\begin{itemize}
    \item $N_j$: number of samples passing through node $j$
    \item $N_{\text{total}}$: total number of samples
    \item $G(t)$: Gini impurity of the parent node
    \item $G(t_{\text{left}}), G(t_{\text{right}})$: Gini impurity of left and right child nodes
\end{itemize}

Finally, normalize the importance scores:

\[
\text{Normalized Importance}(i) = \frac{\text{Importance}(i)}{\sum_{k} \text{Importance}(k)}
\]

\begin{figure}[ht]
    \centering
    \includegraphics[width=\textwidth]{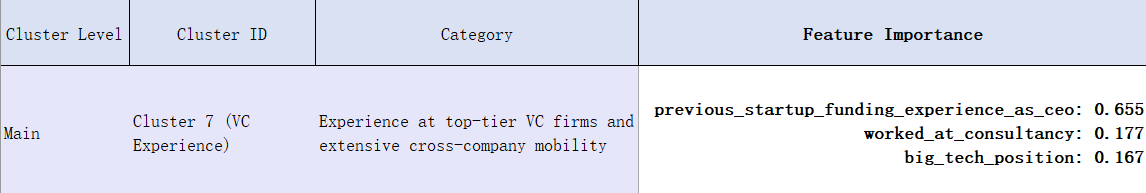}
    \caption{Feature Importance of VC experience cluster}
    \label{fig:feature_importance}
\end{figure}

This normalization ensures that the sum of importance scores across all features equals 1, facilitating comparison of relative importance between different features. For example, as shown in Figure~\ref{fig:feature_importance}, the main contributed features are:
\begin{itemize}
    \item \texttt{previous\_startup\_funding\_as\_ceo}: 0.655,
    \item \texttt{worked\_at\_consultancy}: 0.177,
    \item \texttt{big\_tech\_position}: 0.167.
\end{itemize}

Through this method, we can quantify each feature's contribution within the decision tree, thereby identifying the most influential features in each specific cluster. These feature importance scores not only aid in understanding the model's decision-making process but also provide a basis for subsequent feature selection and model optimization.

\section{Experimental Results}

We evaluate GPT-HTree on Vela Research's proprietary dataset of 8,800 founders with 64 significant features. 

Across all clusters, our analysis revealed several consistently significant features:

\begin{itemize}
    \item \textbf{Previous Experience}: Previous startup funding experience as CEO and management experience in listed companies emerged as the strongest predictors of success.
    
    \item \textbf{Educational Background}: The ranking of the education institution and the level of education showed a substantial impact on the outcomes.
    
    \item \textbf{Professional Network}: Industry connections and professional social media presence demonstrated a significant influence.
\end{itemize}

\subsection{Feature Importance Analysis}

\subsubsection{Global Importance Patterns}
Across all founder clusters, three key dimensions emerged as critical predictors.

\begin{itemize}
    \item \textbf{Startup Track Record}: 
    \begin{itemize}
        \item CEO fundraising experience, especially larger than \$50M
        \item Exit experience through acquisitions
        \item Leadership roles in NASDAQ tech companies
    \end{itemize}
    
    \item \textbf{Professional Experience}: 
    \begin{itemize}
        \item Engineering and PM roles at FAANG companies
        \item Investment roles at VC firms
        \item Startups with tier-1 VC backing
    \end{itemize}
    
    \item \textbf{Public Recognition}:
    \begin{itemize}
        \item Media coverage frequency
        \item Press mentions outside company news
        \item Industry conference speaking
    \end{itemize}
\end{itemize}

\subsubsection{Cluster Success Patterns}
\begin{table}[h!]
\centering
\renewcommand{\arraystretch}{1.2} % Adjust row spacing for compactness
\begin{tabular}{|l|c|p{6cm}|}
\hline
\textbf{Persona} & \textbf{Success Rate} & \textbf{Key Traits} \\ \hline
Serial Exit Founders & 17.4\% & 
Acquired by FAANG companies; Multiple acquisition exits; 15-20 years industry experience \\ \hline

IPO Experience Founders & 16.3\% & 
Led companies to IPOs; Extensive NASDAQ leadership; Strong founding experience \\ \hline

Venture Capitalists & 13.1\% & 
Tier-1 VC experience; Top consulting firm background; Multiple company investments \\ \hline

Tech Leaders & 3.7\% & 
FAANG company experience; Technical leadership roles; NASDAQ company background \\ \hline

Industry Influencers & 2.7\% & 
Significant media presence; Prior startup funding success; Industry achievement records \\ \hline

Professional Athletes & 1.9\% & 
Professional athletic background; Competition achievements; Extracurricular leadership \\ \hline

Career Professionals & 1.4\% & 
Steady career progression; Adaptability indicators; Limited media exposure \\ \hline

Early Professionals & 0.8\% & 
Minimal career experience; Low risk-taking history; Basic educational background \\ \hline
\end{tabular}
\caption{Personas, Success Rates, and Key Traits}
\label{tab:compact_personas}
\end{table}

The table below presents a detailed list of personas, highlighting clear distinctions in success rates driven by factors such as prior exits, industry experience, and leadership roles. Clusters with higher success rates consistently exhibit strong indicators, including acquisition experience, fundraising histories, and senior leadership positions. 

\subsection{Case Studies}

To illustrate the personas, we present real-world case studies below.
\begin{itemize}
\item \textbf{Serial Exit Founder Case}:
\begin{itemize}
\item Previously sold startup to Google for \$200M
\item Led multiple successful exits in enterprise software
\item Strong network in Silicon Valley tech community
\end{itemize}
    \item \textbf{Venture Capitalist Case}:
    \begin{itemize}
        \item Partner at Sequoia Capital for 8 years
        \item Track record of 10+ successful investments
        \item Deep expertise in SaaS business models
    \end{itemize}
    
    \item \textbf{Tech Leader Case}:
    \begin{itemize}
        \item Former Engineering VP at Amazon
        \item Built and scaled major cloud services
        \item Strong technical domain expertise
    \end{itemize}
\end{itemize}

\section{Conclusion}
\textbf{GPT-HTree} addresses the limitations of traditional decision trees by introducing a two-tiered classification approach that first segments the population into homogeneous clusters and then applies localized decision trees. This methodology enhances classification accuracy by ensuring that decision paths are tailored to the unique characteristics of each persona category.

By initially clustering individuals into groups with similar characteristics, GPT-HTree avoids the pitfalls of comparing fundamentally different groups, such as engineers and investors, which can lead to misleading classification results and obscure the true determinants of success within each group. This localized approach ensures that the decision trees are optimized for the specific feature interactions relevant to each persona, thereby providing more accurate and meaningful classifications.

Additionally, the integration of LLMs for persona characterization bridges the gap between quantitative feature analysis and qualitative insight generation, enabling an understanding of the factors driving success within each group. The identification of key features within clusters provides actionable insights that can inform targeted strategies for startup investments.

\subsection{Limitations}
False positives in our analysis often occurred due to an over-emphasis on superficial factors, such as media presence without substantive achievements, inflated valuation histories, and surface-level industry connections. Conversely, false negatives were primarily associated with undervalued domain expertise in emerging sectors and overlooking first-time founders. These challenges are further compounded by certain limitations in the model itself, including a reliance on historical patterns, bias toward traditional success metrics, and a limited ability to analyze market characteristics such as timing. 

Additionally, this study was constrained by the capabilities of GPT-4o. The dataset also contains inherent biases, including the possibility of founders omitting or exaggerating information in their profiles. Furthermore, data quality issues might exist due to the LLM hallucinations during the feature engineering phase of data preparation. 

Finally, it's important to note that this paper is intended solely for research purposes and should not be interpreted as financial or investment advice.

\subsection{Future Work}

Below, we outline key areas for future exploration and development.

\begin{itemize} 

    \item \textbf{Extensibility to other domains:} GPTHTree can be applied to other fields involving high-stakes decision-making, such as healthcare, while offering accuracy, interpretability, and actionable insights. 

    \item \textbf{Multi-modal applications:} The model can be extended to multi-modal use cases, including images, audio, and video. 

    \item \textbf{Feature engineering and success criteria:} Instead of using resampling methods, success criteria can be adjusted to increase the number of success cases. Additional features can also be included to group teams and markets more effectively. 
    
    \item \textbf{Exploration of other hybrid models:} This research focuses on combining LLMs and ML models. Future work could explore other clustering and classification models. 
    
    \item \textbf{Incorporating more advanced LLMs:} New, more advanced models are continually being developed, such as OpenAI o1 and o3 models, which could be evaluated in future research. 
    
    \item \textbf{Improved prompting techniques:} Techniques like prompt chaining, self-play, and reflexion can be explored to enhance the quality of persona descriptions. 
    
    \item \textbf{Parameter tuning:} The process of determining the number of clusters and subclusters could be improved further using visualization techniques. 
    
\end{itemize}

\bibliographystyle{ieeetr}
\bibliography{references}

\begin{thebibliography}{10}

\bibitem{Xiong2024}
S.~Xiong, Y.~Ihlamur, F.~Alican, and Y.~A., ``Gptree: Towards explainable decision-making via llm-powered decision trees,'' {\em arXiv}, 2024.

\bibitem{gu2024large}
Y.~Gu, H.~You, J.~Cao, and M.~Yu, ``Large language models for constructing and optimizing machine learning workflows: A survey,'' {\em arXiv preprint arXiv:2411.10478}, 2024.

\bibitem{Duda2001}
R.~Duda, P.~Hart, and D.~Stork, {\em Pattern Classification}.
\newblock New York: Wiley, 2~ed., 2001.

\bibitem{Durkin1992}
J.~Durkin, ``Induction via id3,'' {\em AI Expert}, 1992.

\bibitem{Fayyad1992}
U.~M. Fayyad and K.~B. Irani, ``On the handling of continuous-values attributes in decision tree generation,'' {\em Machine Learning}, 1992.

\bibitem{Breiman1993}
L.~Breiman, J.~Friedman, R.~Olshen, and C.~Stone, {\em Classification and Regression Trees}.
\newblock New York: Chapman Hall, 1993.

\bibitem{Quinlan1993}
J.~Quinlan, {\em Programs for Machine Learning}.
\newblock San Francisco: Morgan Kaufmann, 1993.

\bibitem{Quinlan1996}
J.~Quinlan, ``Improved use of continuous attributes in c4.5,'' {\em Journal of Artificial Intelligence}, 1996.

\bibitem{Brodley1995}
C.~Brodley and P.~Utgoff, ``Multivariate decision trees,'' {\em Machine Learning}, 1995.

\bibitem{Yang1997}
Y.~Yang and J.~Pedersen, ``A comparative study on feature selection in text categorization,'' in {\em Proceedings of the 14th International Conference on Machine Learning (ICML'97)}, 1997.

\bibitem{Riley1989}
M.~Riley, ``Some applications of tree based modeling to speech and language indexing,'' 1989.

\bibitem{Chien2002}
J.~Chien, C.~Huang, and S.~Chen, ``Compact decision trees with cluster validity for speech recognition,'' in {\em Proceedings of the IEEE International Conference on Acoustics, Speech, and Signal Processing}, 2002.

\bibitem{Salzberg1998}
S.~Salzberg, A.~Delcher, K.~Fasman, and J.~Henderson, ``A decision tree system for finding genes in dna,'' {\em Journal of Computational Biology}, 1998.

\bibitem{Zamir1998}
O.~Zamir and O.~Etzioni, ``Web document clustering: A feasibility demonstration,'' in {\em Research and Development in Information Retrieval}, 1998.

\bibitem{Cho2002}
Y.~Cho, J.~Kim, and S.~Kim, ``A personalized recommender system based on web usage mining and decision tree induction,'' {\em Expert Systems with Applications}, 2002.

\bibitem{Hollmann2024}
N.~Hollmann, S.~M{\"u}ller, and F.~Hutter, ``Large language models for automated data science: Introducing caafe for context-aware automated feature engineering,'' in {\em Advances in Neural Information Processing Systems}, 2024.

\bibitem{Wang2024a}
B.~Wang, Z.~Wang, X.~Wang, Y.~Cao, R.~A. Saurous, and Y.~Kim, ``Grammar prompting for domain-specific language generation with large language models,'' in {\em Advances in Neural Information Processing Systems}, 2024.

\bibitem{Achiam2023}
J.~Achiam, S.~Adler, S.~Agarwal, L.~Ahmad, I.~Akkaya, F.~L. Aleman, D.~Almeida, J.~Altenschmidt, S.~Altman, S.~Anadkat, {\em et~al.}, ``Gpt-4 technical report,'' {\em arXiv preprint arXiv:2303.08774}, 2023.

\bibitem{Touvron2023}
H.~Touvron, T.~Lavril, G.~Izacard, X.~Martinet, M.-A. Lachaux, T.~Lacroix, B.~Rozi{\`e}re, N.~Goyal, E.~Hambro, F.~Azhar, {\em et~al.}, ``Llama: Open and efficient foundation language models,'' {\em arXiv preprint arXiv:2302.13971}, 2023.

\bibitem{Hu2024}
W.~Hu, Y.~Xu, Y.~Li, W.~Li, Z.~Chen, and Z.~Tu, ``Bliva: A simple multimodal llm for better handling of text-rich visual questions,'' in {\em Proceedings of the AAAI Conference on Artificial Intelligence}, 2024.

\bibitem{Tai2024}
Y.~Tai, W.~Fan, Z.~Zhang, and Z.~Liu, ``Link-context learning for multimodal llms,'' in {\em Proceedings of the IEEE/CVF Conference on Computer Vision and Pattern Recognition}, 2024.

\bibitem{Luo2024}
D.~Luo, C.~Feng, Y.~Nong, and Y.~Shen, ``Autom3l: An automated multimodal machine learning framework with large language models,'' {\em arXiv preprint arXiv:2408.00665}, 2024.

\bibitem{Gu2023}
Y.~Gu, J.~Cao, Y.~Guo, S.~Qian, and W.~Guan, ``Plan, generate and match: Scientific workflow recommendation with large language models,'' in {\em International Conference on Service-Oriented Computing}, Springer, 2023.

\bibitem{Klievtsova2023}
N.~Klievtsova, J.-V. Benzin, T.~Kampik, J.~Mangler, and S.~Rinderle-Ma, ``Conversational process modelling: state of the art, applications, and implications in practice,'' in {\em International Conference on Business Process Management}, Springer, 2023.

\bibitem{Zhang2023a}
Z.~Zhang, A.~Zhang, M.~Li, H.~Zhao, G.~Karypis, and A.~Smola, ``Multimodal chain-of-thought reasoning in language models,'' {\em arXiv preprint arXiv:2302.00923}, 2023.

\bibitem{Xiao2024}
T.~Z. Xiao, R.~Bamler, B.~Sch{\"o}lkopf, and W.~Liu, ``Verbalized machine learning: Revisiting machine learning with language models,'' {\em arXiv preprint arXiv:2406.04344}, 2024.

\bibitem{Hong2024}
S.~Hong, Y.~Lin, B.~Liu, B.~Wu, D.~Li, J.~Chen, J.~Zhang, J.~Wang, L.~Zhang, M.~Zhuge, {\em et~al.}, ``Data interpreter: An llm agent for data science,'' {\em arXiv preprint arXiv:2402.18679}, 2024.

\bibitem{Lazebnik2022}
T.~Lazebnik, A.~Somech, and A.~I. Weinberg, ``Substrat: A subset-based optimization strategy for faster automl,'' {\em Proceedings of the VLDB Endowment}, 2022.

\bibitem{Nikitin2022}
N.~O. Nikitin, P.~Vychuzhanin, M.~Sarafanov, I.~S. Polonskaia, I.~Revin, I.~V. Barabanova, G.~Maximov, A.~V. Kalyuzhnaya, and A.~Boukhanovsky, ``Automated evolutionary approach for the design of composite machine learning pipelines,'' {\em Future Generation Computer Systems}, 2022.

\bibitem{An2023}
S.~An, H.~Lee, J.~Jo, S.~Lee, and S.~J. Hwang, ``Diffusionnag: Predictor-guided neural architecture generation with diffusion models,'' in {\em The Twelfth International Conference on Learning Representations}, 2023.

\bibitem{Dakhel2023}
A.~M. Dakhel, V.~Majdinasab, A.~Nikanjam, F.~Khomh, M.~C. Desmarais, and Z.~M.~J. Jiang, ``Github copilot ai pair programmer: Asset or liability?,'' 2023.

\bibitem{ArteagaGarcia2024}
E.~J. Arteaga~Garcia, J.~F.~N. Pimentel, Z.~Feng, M.~Gerosa, I.~Steinmacher, and A.~Sarma, ``How to support ml end-user programmers through a conversational agent,'' in {\em Proceedings of the 46th IEEE/ACM International Conference on Software Engineering}, 2024.

\bibitem{Zhang2023b}
L.~Zhang, Y.~Zhang, K.~Ren, D.~Li, and Y.~Yang, ``Mlcopilot: Unleashing the power of large language models in solving machine learning tasks,'' {\em arXiv preprint arXiv:2304.14979}, 2023.

\bibitem{Nam2024}
J.~Nam, K.~Kim, S.~Oh, J.~Tack, J.~Kim, and J.~Shin, ``Optimized feature generation for tabular data via llms with decision tree reasoning,'' {\em arXiv preprint arXiv:2406.08527}, 2024.

\bibitem{Zhang2024a}
X.~Zhang, J.~Zhang, B.~Rekabdar, Y.~Zhou, P.~Wang, and K.~Liu, ``Dynamic and adaptive feature generation with llm,'' {\em arXiv preprint arXiv:2406.03505}, 2024.

\bibitem{Bommasani2021}
R.~Bommasani, D.~A. Hudson, E.~Adeli, R.~Altman, S.~Arora, S.~von Arx, M.~S. Bernstein, J.~Bohg, A.~Bosselut, E.~Brunskill, {\em et~al.}, ``On the opportunities and risks of foundation models,'' {\em arXiv preprint arXiv:2108.07258}, 2021.

\bibitem{Yao2024a}
Y.~Yao, J.~Duan, K.~Xu, Y.~Cai, Z.~Sun, and Y.~Zhang, ``A survey on large language model (llm) security and privacy: The good, the bad, and the ugly,'' {\em High-Confidence Computing}, 2024.

\bibitem{xu2019modeling}
L.~X. M. S.~A. C. and K.~V., ``Modeling tabular data using conditional gan,'' 2019.

\end{thebibliography}

\end{document}